\documentclass{article}
\usepackage[preprint]{colm2026_conference}

\usepackage{microtype}
\usepackage{hyperref}
\usepackage{url}
\usepackage{booktabs}
\usepackage{lineno}
\usepackage{graphicx}
\usepackage{amsmath}
\usepackage{amssymb}
\usepackage{mathtools}
\usepackage[capitalize,noabbrev]{cleveref}

\definecolor{darkblue}{rgb}{0, 0, 0.5}
\hypersetup{colorlinks=true, citecolor=darkblue, linkcolor=darkblue, urlcolor=darkblue}

\newcommand{\sarvam}{Sarvam-30B}
\newcommand{\qwenmoe}{Qwen3\mbox{-}30B\mbox{-}A3B}
\newcommand{\vharm}{\hat{v}_{\mathrm{harm}}}
\newcommand{\Cohen}{\textit{d}}

\title{Decided Upstream, Written Late:\\
       Locating and Pricing the Cross-Lingual Refusal Circuit of a Multilingual MoE}

\author{Ramakrishna P. Kompella \\
  G\"odel Machines \\
  \texttt{rk@goedelmachines.com} \\
  \And
  Aadit Mahajan \\
  G\"odel Machines \\
  IIT Madras \\
  \texttt{aadit2003@outlook.com} \\
}

\begin{document}

\ifcolmsubmission
\linenumbers
\fi

\maketitle

% [preprint] sets "Preprint. Under review."; [final] would claim the main conference.
% Neither fits a workshop paper. String below matches the workshop's house style
% (cf. arXiv:2607.29062, accepted at the same workshop).
\lhead{Published at the Workshop on Actionable Interpretability at COLM 2026}

\begin{abstract}
Safety alignment in multilingual models is uneven: a model that reliably refuses a
harmful request in English will often comply with the same request in a lower-resource
language. We trace this gap mechanistically in \sarvam{}, an Indic-multilingual
mixture-of-experts reasoning model, and find it is not a failure to \emph{detect} harm.
Harm is encoded as an internal direction that is nearly language-invariant in
mid-network (English-vs-Indic cosine ${\approx}0.9$ at $L11$), and steering that
direction upstream causally controls refusal. But the detection direction is
\emph{orthogonal} to the change that actually \emph{writes} the refusal, which is late
and assembled over the course of generation rather than read off in a single forward
pass. We attribute the write to a specific, localizable circuit---a mixture-of-experts
\emph{writer} held in check by an attention \emph{opposer}---and price every way of
intervening on it: damping the opposer is cheap and effective, amplifying the writer is
a cost wall, and surgical edits to the responsible heads do nothing. The circuit's
organization, and the gradient method that exposes it, recur in a second, unrelated
MoE model, while the lever's strength is architecture-specific. The result is a
cost-measured map of where a multilingual safety repair can land, and what it costs.
\end{abstract}

\section{Introduction}\label{sec:intro}
Large language models are trained to refuse a class of malicious requests rather than help 
with them. Making a model refuse reliably is a central goal of safety alignment, and in 
frontier English models it works well enough to seem simply ``built in.''

However, it is not built in evenly. A model that refuses a request in English will often comply
when the same request is posed in a lower-resource language; a \emph{cross-lingual
safety gap} that is a real vulnerability for multilingual deployments \citep{Wang2026-iq},
and most acute for models built around lower-resource languages, such as \sarvam{}
\citep{sarvam2025}, an open reasoning model for English and the major Indic languages.

Why a model refuses is increasingly something we can inspect directly rather than treat
as a black box. Mechanistic interpretability has shown that refusal can often be captured
by a single direction in a model's internal activations; in mixture-of-experts
(MoE) models, which route each token through a few of many specialized \emph{experts}, it
can concentrate in particular experts \citep{Lai2025-yv}. Located precisely, the
machinery of safety can be measured, compared across languages, and repaired at its
source.

These accounts come almost entirely from English, dense models. Whether they describe a
multilingual MoE \emph{reasoning} model, and whether they explain the cross-lingual gap,
is open. We investigate the mechanisms of refusal and safety in \sarvam{}, tracing the
behaviour from where the model registers harm to where a refusal is produced. We also test
how far the account generalizes on a second multilingual MoE.

\section{Setup}\label{sec:setup}
\textbf{Model.} \sarvam{} \citep{sarvam2025} is a 30B-parameter, 19-layer
($L0$ dense, $L1$--$18$ MoE, 128 experts, top-6) Indic-multilingual reasoning model
that emits an explicit \texttt{<think>} trace. We use six languages (English plus
Hindi, Bengali, Punjabi, Tamil, Malayalam).
\textbf{Data/labels.} The paired-prompt TwinBreak corpus \citep{Wang2026-iq,309742}
pairs each malicious prompt with a matched benign twin, translated into the six
languages; generations are labelled \textsc{refuse}/\textsc{comply} by a
broadened-regex pre-pass and an LLM judge (agreement $38/39$, \cref{app:repro}).
``Recovery'' is the fraction of \emph{compliant-malicious} prompts an intervention
flips to a clean \textsc{refuse}.
\textbf{Directions.} The per-(language, layer) harm axis $\vharm$ is the input-position
difference-of-means of harmful vs.\ benign prompts \citep{Arditi2024-zd}; the refusal
direction is the verdict-position difference-of-means of refused vs.\ complied
generations, constructed independently of $\vharm$. \textbf{Cost.} Fluency cost is
WikiText-2 perplexity over $64\times512$-token chunks; the base model scores
${\approx}15$.

\section{Harm is detected, shared across languages, and orthogonal to refusal}\label{sec:detect}

\paragraph{Detection lives in a language-agnostic mid-network band.} A logistic
language probe over the residual stream falls from $99.6\%$ at $L1$ to $40\%$ at $L11$
and recovers to $96\%$ by $L18$ (chance $17\%$): the middle of the network strips
surface language. That same band is where harm becomes linearly decodable, and the harm
axis there is cross-lingually \emph{shared}---per-language harm axes align at
$0.95\pm0.01$ Indic$\leftrightarrow$Indic and $0.90\pm0.01$
English$\leftrightarrow$Indic at $L11$ (\cref{app:detect}). English's encoding is
measurably stronger (axis norm $37.4$ vs.\ $22$--$25$ for any Indic language;
${\sim}7$pp higher probe accuracy), a quantitative head-start that foreshadows the gap.

\begin{figure}[t]
\centering
\includegraphics[width=\linewidth]{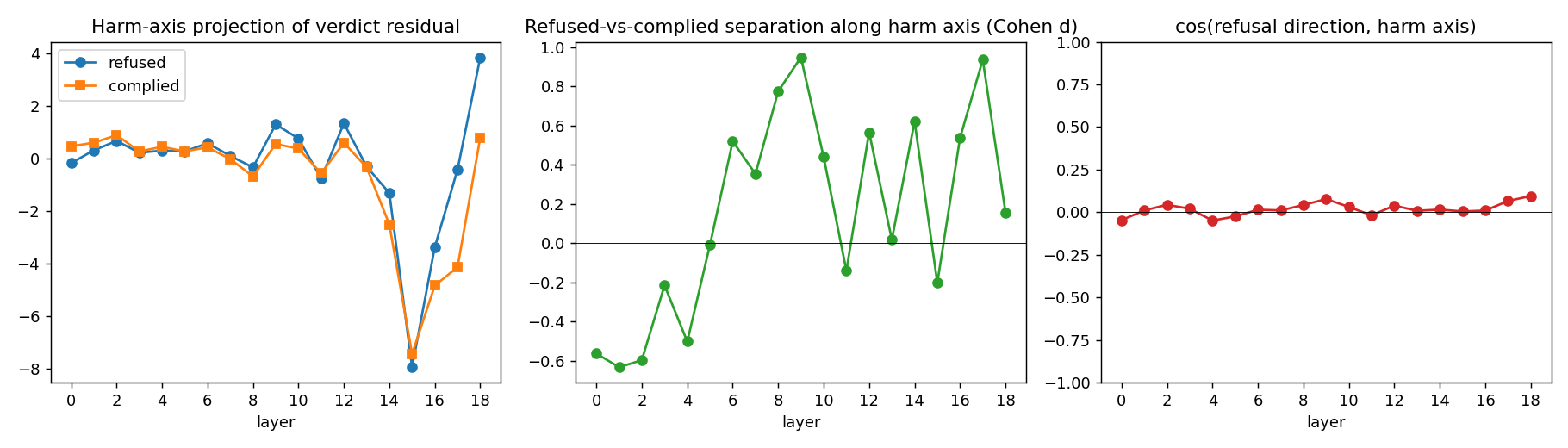}
\caption{\textbf{The behaviour-writing change is late and orthogonal to detection.}
\emph{Left}: harm-axis projection of the verdict residual, refused vs.\ complied.
\emph{Middle}: refused$-$complied separation along the harm axis (Cohen $\Cohen$)---a
weak mid-layer \emph{detection} signal. \emph{Right}: $\cos$(refusal direction, harm
axis) $\approx0$ at every layer---the large late residual movement that \emph{is} the
refusal is orthogonal to the detection axis.}
\label{fig:subspaces}
\end{figure}

\paragraph{But detection is orthogonal to the change that writes refusal.} We ask, per
layer, (i) whether refused and complied generations \emph{separate} along the harm
axis, and (ii) whether the residual \emph{movement} when refusing aligns with that axis
(\cref{fig:subspaces}). They dissociate. Harm \emph{separation} is a weak, middle-layer
signal (Cohen $\Cohen$ peaks at $0.95$ at $L9$). The residual change that accompanies
refusal is near-\emph{orthogonal} to the harm axis at every layer ($|\cos|\le0.10$, max
$0.095$ at $L18$; per-language cosines $0.04$--$0.15$, \cref{app:detect}) and its
\emph{magnitude} grows monotonically into the late layers
($\lVert\Delta\rVert: 2.2$ at $L4 \to 13.7$ at $L11 \to 81.7$ at $L18$). So
``is this harmful?'' (mid, weak, $\parallel\vharm$) and ``write the refusal'' (late,
large, $\perp\vharm$) are different directions in different layers
\citep{Zhao2025-fn}. The behaviour lives in a \emph{late} subspace, separate from
detection---which is why uniform detection and uneven refusal can coexist.

\section{Refusal write is late and carried by generation}\label{sec:gen}

\paragraph{Single forward pass cannot reach the late subspace.} If execution were a
feed-forward read-out of detection, injecting the detection direction upstream would
drive the output. It does not (\cref{tab:prop}). Adding $\alpha\vharm$ at $L10$ lifts
the refusal-token log-prob \emph{locally} ($+1.7$ nats at $L10$), but the lift decays
within ${\sim}2$ layers and is \emph{gone} at the output: at $L18$ the steered log-prob
($-39.7$) is no better than unsteered ($-38.4$). A layer sweep confirms the result is
general---steering at $L15$ alone is a no-op, and multi-layer injection extends the
lifted band by only ${\sim}1$--$2$ layers; \emph{no} single-pass configuration moves
the $L15$--$18$ band.

\paragraph{Re-applying detection at every step does.} Activation addition of
$\vharm$ at $L10$, \emph{re-applied at every decode step}, recovers refusal with recovery
$0.65\,[0.50,0.78]$, $16\times$ malicious/benign selectivity, and $4\%$ benign knockout;
a layer sweep places the causal ceiling at $L11$--$12$ (\cref{app:steer}). This is the
positive counterpart to \cref{tab:prop}: the same direction that \emph{cannot} reach
the output in one pass \emph{does} produce refusal when carried across generation.
Refusal is therefore not a single-pass feed-forward read-out; the decision is realized
over the autoregressive trajectory. This is what makes the cost structure of
\cref{sec:circuit} the way it is: the affordable interventions \emph{scale or release}
a component the model already produces in distribution (the upstream detection signal,
or the opposer's brake), whereas writing refusal at the output means driving late
activations out of distribution.

\begin{table}[t]
\centering
\small
\caption{\textbf{Single-pass injection dies before the output.} Refusal-token log-prob
(logit lens) by layer after adding $\vharm$ at $L10$, vs.\ unsteered. The local lift at
$L10$ decays within ${\sim}2$ layers and is absent at the output ($L18$). Multi-layer
injection likewise never reaches $L15$--$18$.}
\label{tab:prop}
\begin{tabular}{lcccc}
\toprule
Layer & 10 & 12 & 14 & 18 (out) \\
\midrule
unsteered        & $-6.05$ & $-10.60$ & $-17.51$ & $-38.39$ \\
$+\vharm$ @ $L10$ & $-4.35$ & $-8.98$  & $-16.89$ & $-39.70$ \\
lift (nats)      & $+1.70$ & $+1.62$  & $+0.62$  & $-1.31$ \\
\bottomrule
\end{tabular}
\end{table}

\section{Attributing the late write}\label{sec:attr}
\begin{figure}[t]
\centering
\includegraphics[width=\linewidth]{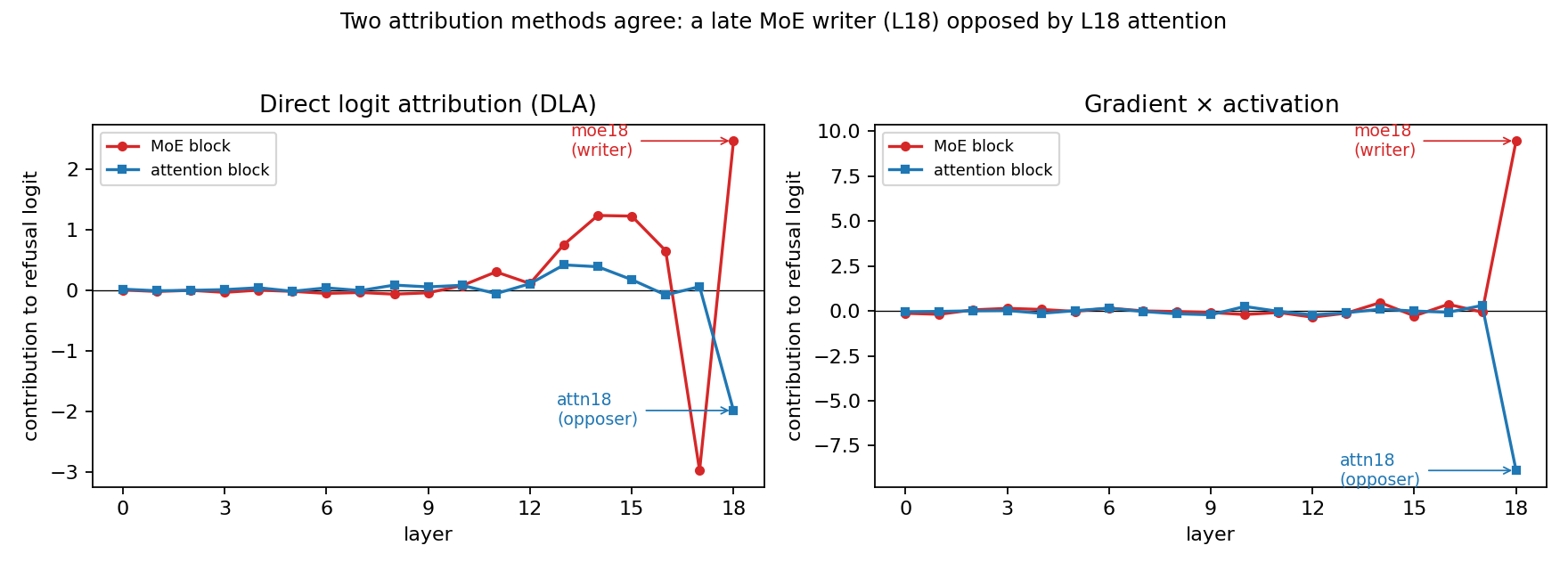}
\caption{\textbf{Two attribution methods agree on a late writer/opposer.} Per-layer
contribution to the refusal logit, MoE block vs.\ attention block, averaged over
refusers. \emph{Left}: direct logit attribution. \emph{Right}:
gradient$\times$activation. Both put a dominant positive write at \emph{moe18} and a
dominant negative term at \emph{attn18}.}
\label{fig:attr}
\end{figure}

What writes the late, orthogonal change? We use two independent attribution methods on
the refusers (\cref{fig:attr}). \textbf{Direct logit attribution} (DLA;
\citealp{nostalgebraist2020}) decomposes the refusal-token logit into per-block
contributions: the MoE blocks build a positive (pro-refusal) contribution across
$L14$--$16$, and at $L17$--$18$ the decomposition splits into a large \emph{negative}
$moe17$ term ($-3.0$) and a positive $moe18$ write ($+2.5$) of comparable
magnitude---a sign pattern characteristic of logit-lens decompositions near the
output---while $L18$ attention is strongly negative (\emph{attn18}, $-2.0$). Because
these raw DLA terms partially cancel, we anchor the conclusion on
\textbf{gradient$\times$activation} total-effect attribution, which localizes the net
effect sharply to $L18$ (\emph{moe18} $+9.5$, \emph{attn18} $-8.9$; all other layers
near zero) and agrees with DLA on sign and location. The two methods---a linear logit
decomposition and a first-order causal estimate---thus converge: a late MoE
\emph{writer} opposed by late attention. At the expert level the $L18$ shared expert and
a handful of $L15$--$16$ experts carry the writer term; for the opposer, the gradient
and projection-based head rankings overlap on $L18$ heads $53$, $61$, $17$, and $27$.

\section{Pricing the circuit, and what transfers}\label{sec:circuit}
The attribution gives a small, manipulable circuit: MoE \emph{writers} ($L18$, plus
specific $L15$--$16$ experts and the $L18$ shared expert) push toward refusal, and an
$L18$ attention \emph{opposer} pushes against it, so damping it \emph{raises} refusal.
The circuit is also language-structured---consistent with detection being shared but
execution being language-specific, the top-$K$ harm-expert routing Jaccard between Indic
languages and English collapses from $0.43$ at $L11$ to $0.03$ at $L18$
(\cref{app:moe}), a candidate mechanism for the gap itself. Because the circuit exposes
three handles---amplify writers, damp the opposer, or edit heads surgically---we price
each by intervening and measuring refusal recovery against WikiText perplexity
(\cref{tab:map}).

\paragraph{The asymmetry.} \emph{Damping the opposer is cheap and effective.}
\texttt{attn18}$\times0.5$ recovers $0.54$ of compliant-malicious prompts at
$1.09\times$ perplexity with near-zero benign collateral ($1/35$): you do not
\emph{add} refusal, you remove the brake. \emph{Writing} refusal is a cost wall:
\texttt{moe18}$\times2$ costs $2.2\times$ perplexity for partial, degraded suppression,
and \texttt{shared18}$\times3$ costs $24\times$ and collapses fluency into
\texttt{</think>}-spam---pushing the writer harder drives late activations out of
distribution, exactly as \cref{sec:gen} predicts. \emph{Surgical head edits are precise,
cheap, and inert}: rank-restricted edits and ablations of the attributed heads leave
perplexity at base ($0.94$--$1.01\times$) but induce no refusal above base (judged
\textsc{refuse} $0$--$6\%$, base $3\%$). The actionable content is the asymmetry: the
affordable execution-side repair is \emph{suppressive} (damp the opposer), not
generative (amplify the writer)---and upstream, seeding detection recovers refusal at
$0.65$ (\cref{sec:gen}).

\begin{table}[t]
\centering
\small
\caption{\textbf{Cost map (selected sites).} Recovery = compliant-malicious flipped to
\textsc{refuse}; ppl = WikiText perplexity ratio to base. Upstream detection-steering
($0.65$ recovery, \cref{sec:gen}) is omitted: its perplexity is unmeasured.}
\label{tab:map}
\begin{tabular}{llcc}
\toprule
Family & Site & Recov. & ppl$\times$ \\
\midrule
damp (opposer)     & \texttt{attn18}$\times0.5$ & $0.54$ & $1.09$ \\
damp (opposer)     & \texttt{attn18}$\times0.25$& $0.37$ & $1.92$ \\
write (amplify)    & \texttt{moe18}$\times2$    & $0.50^{\dagger}$ & $2.16$ \\
write (amplify)    & \texttt{shared18}$\times3$ & $0.00$ & $23.7$ \\
surgical head      & \texttt{dir-top8}          & $0.03^{\S}$ & $1.01$ \\
surgical head      & \texttt{ablate-top4}       & $0.00^{\S}$ & $0.94$ \\
\bottomrule
\end{tabular}
\\[2pt]
{\footnotesize $^{\dagger}$ small dose set ($n{=}6$); outputs degraded, not clean refusals.
$^{\S}$ LLM-judged (same rubric); equals base ($0.03$), so inert---but most outputs did not
commit within the generation budget (a weaker test than the run-to-EOS damp comparison).}
\end{table}

\paragraph{Cross-model: the organization transfers, the lever does not.} We re-ran the
pipeline on a second, unrelated MoE---\qwenmoe{} \citep{qwen3technicalreport}. The
organization replicates. The cross-lingual safety gap recurs (\qwenmoe{} complies with
$0.28$ of Indic-mean harmful prompts vs.\ $0.07$ in English); a late MoE block
\emph{writes} refusal by gradient attribution ($+10.2$ at $L47$); and the \emph{attribution
lesson} transfers exactly---DLA shows near-zero attention, but gradient$\times$activation
reveals an \emph{opposer of equal magnitude} ($L40$, $-10.2$ against the writer's $+10.2$),
just as it surfaced \texttt{attn18} in \sarvam{}. The cost signature transfers too: damping
the opposer is near-free ($1.00\times$ perplexity) while amplifying the writer costs fluency
($1.20\times$). What does \emph{not} transfer is the lever's strength: amplifying the writer
recovers refusal only partially---the best whole-band amplification adds $+24$pp (to $59\%$
refusal) and then \emph{saturates}, shattering the model past $\times3$. The governing factor
is whether the model has a refusal-specific, always-on component to amplify cleanly:
\sarvam{}'s shared expert tips behaviour with low collateral, whereas \qwenmoe{}'s aggregate
routed block scales everything at once, so it garbles before it fully flips. The map and the
methodology transfer; how far the lever moves behaviour is a property of the architecture.

\section{Discussion}\label{sec:disc}
The cross-lingual safety gap is not a detection failure. \sarvam{} represents harm in a
shared, near language-invariant mid-network direction, and that direction causally
controls refusal when steered upstream. The gap appears \emph{downstream}: the direction
a single forward pass can read off (detection) is not the direction that produces the
behaviour (a late, orthogonal, MoE-written change), the link between them is the decode
trajectory, and the late expert routing that turns the shared signal into a refusal
token becomes increasingly language-specific (Jaccard $0.43{\to}0.03$). For actionable
interpretability this reframes ``what is the refusal feature?'' into ``where, and at
what cost, can I intervene on a computation that unfolds over generation?''---with a
clean answer: seed detection upstream, or suppress the late opposer; do not try to write
refusal at the output. The attribution methodology and the writer/opposer organization
transfer to \qwenmoe{}; the affordances (a sharp, cheap opposer; the orthogonal
geometry) are model-specific and worth measuring per model before editing.

\section{Limitations}\label{sec:lim}
One feature (refuse/comply) characterized in depth in one model, plus a cross-model
check. (i) \sarvam{} is a \emph{reasoning} model; the generation-time and orthogonality
account may differ without a \texttt{<think>} trace. Our cross-model check (\qwenmoe{}) is
\emph{non-reasoning}, so it tests the attribution and lever transfer but not the
generation-time claims, which remain \sarvam{}-only; a faithful test needs a
thinking-mode MoE (e.g.\ Qwen3-Thinking), which we did not run. (ii) The
writer-amplification recovery is from a small dose set ($n{=}6$) and is noisy.
(iii) Recovery and harm labels rely on an LLM judge; we report regex agreement but not
human adjudication at scale. (iv) The cost map is indicative, not strictly
apples-to-apples: recovery is measured on heterogeneous eval sets, and upstream
detection-steering recovers refusal but its perplexity was not measured, so it is
omitted from the quantitative map. (v) All interventions studied \emph{increase}
refusal; the cost asymmetry is offered to guide where defensive alignment edits are
affordable, not to enable the reverse.

\bibliographystyle{colm2026_conference}
\bibliography{refs}

@ARTICLE{Arditi2024-zd,
  title         = "Refusal in language models is mediated by a single direction",
  author        = "Arditi, Andy and Obeso, Oscar and Syed, Aaquib and Paleka,
                   Daniel and Panickssery, Nina and Gurnee, Wes and Nanda, Neel",
  month         =  jun,
  year          =  2024,
  copyright     = "http://arxiv.org/licenses/nonexclusive-distrib/1.0/",
  journal       = "arXiv preprint arXiv:2406.11717",
  archivePrefix = "arXiv",
  primaryClass  = "cs.LG",
  eprint        = "2406.11717"
}

@ARTICLE{Zhao2025-fn,
  title         = "{LLMs} encode harmfulness and refusal separately",
  author        = "Zhao, Jiachen and Huang, Jing and Wu, Zhengxuan and Bau,
                   David and Shi, Weiyan",
  month         =  dec,
  year          =  2025,
  copyright     = "http://creativecommons.org/licenses/by/4.0/",
  journal       = "arXiv preprint arXiv:2507.11878",
  archivePrefix = "arXiv",
  primaryClass  = "cs.CL",
  eprint        = "2507.11878"
}

@misc{sarvam2025,
  title         = {Sarvam-30B: An Open Indic-Multilingual Mixture-of-Experts Reasoning Model},
  author        = {{Sarvam AI}},
  year          = {2025},
  url           = {https://huggingface.co/sarvamai/sarvam-30b}
}

@misc{nostalgebraist2020,
  title         = {Interpreting GPT: the logit lens},
  author        = {nostalgebraist},
  year          = {2020},
  howpublished  = {LessWrong},
  url           = {https://www.lesswrong.com/posts/AcKRB8wDpdaN6v6ru/interpreting-gpt-the-logit-lens}
}

@inproceedings {309742,
author = {Torsten Krau{\ss} and Hamid Dashtbani and Alexandra Dmitrienko},
title = {{TwinBreak}: Jailbreaking {LLM} Security Alignments based on Twin Prompts},
booktitle = {34th USENIX Security Symposium (USENIX Security 25)},
year = {2025},
isbn = {978-1-939133-52-6},
address = {Seattle, WA},
pages = {2343--2362},
url = {https://www.usenix.org/conference/usenixsecurity25/presentation/krauss},
publisher = {USENIX Association},
month = aug
}

@ARTICLE{Wang2026-iq,
  title         = "Refusal direction is universal across safety-aligned
                   languages",
  author        = "Wang, Xinpeng and Wang, Mingyang and Liu, Yihong and
                   Sch{\"u}tze, Hinrich and Plank, Barbara",
  month         =  feb,
  year          =  2026,
  copyright     = "http://creativecommons.org/licenses/by/4.0/",
  journal       = "arXiv preprint arXiv:2505.17306",
  archivePrefix = "arXiv",
  primaryClass  = "cs.CL",
  eprint        = "2505.17306"
}

@ARTICLE{Lai2025-yv,
  title         = "{SAFEx}: Analyzing vulnerabilities of {MoE-based} {LLMs} via
                   stable safety-critical expert identification",
  author        = "Lai, Zhenglin and Liao, Mengyao and Wu, Bingzhe and Xu, Dong
                   and Zhao, Zebin and Yuan, Zhihang and Fan, Chao and Li,
                   Jianqiang",
  month         =  oct,
  year          =  2025,
  copyright     = "http://arxiv.org/licenses/nonexclusive-distrib/1.0/",
  journal       = "arXiv preprint arXiv:2506.17368",
  archivePrefix = "arXiv",
  primaryClass  = "cs.LG",
  eprint        = "2506.17368"
}

@misc{qwen3technicalreport,
      title={Qwen3 Technical Report}, 
      author={QwenTeam},
      year={2025},
      eprint={2505.09388},
      archivePrefix={arXiv},
      primaryClass={cs.CL},
      url={https://arxiv.org/abs/2505.09388}, 
}

\appendix
\section{Cross-lingual detection detail}\label{app:detect}
\begin{figure}[h]
\centering
\includegraphics[width=0.7\linewidth]{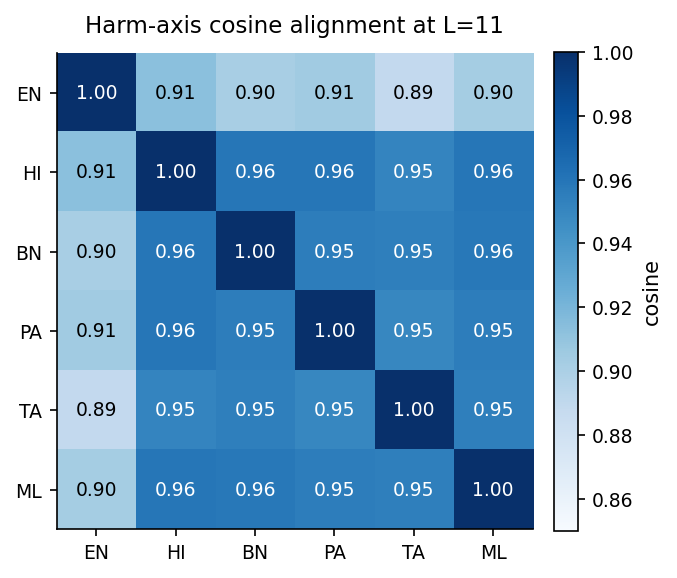}
\caption{Cross-lingual cosine alignment of input-position harm axes at $L11$ (bootstrap
means; std $\le0.014$). Indic$\leftrightarrow$Indic cluster at $0.95\pm0.01$;
English$\leftrightarrow$Indic at $0.90\pm0.01$.}
\label{fig:cosine}
\end{figure}

Per-language input-position harm axes are diff-of-means with bootstrap CIs over 500
paired resamples on the 100 (malicious, benign) twins \citep{Arditi2024-zd}.
Cross-lingual cosine alignment at $L11$ (\cref{fig:cosine}): Indic$\leftrightarrow$Indic
$0.95\pm0.01$, English$\leftrightarrow$Indic $0.90\pm0.01$. The input-position harm axis
and the output-position refusal axis are orthogonal in every language at $L11$
(EN $0.04$, HI $0.06$, BN $0.06$, PA $0.11$, TA $0.15$, ML $0.10$) and remain so through
$L0$--$14$; weak alignment appears only at the residual-norm-outlier layer $L18$,
replicating \citet{Zhao2025-fn} in a non-English MoE setting. During Indic generation
the model deliberates in English-script tokens inside \texttt{<think>} (ASCII fraction
$0.90$--$0.95$), and a logit lens shows the residual stream framing its decision in
English safety vocabulary (\textit{unfortunately}, \textit{harmful}) by $L9$, consistent
with \citet{Wang2026-iq}'s universal cross-lingual refusal direction.

\section{Upstream steering: dose-response and layer sweep}\label{app:steer}
\begin{figure}[h]
\centering
\includegraphics[width=0.49\linewidth]{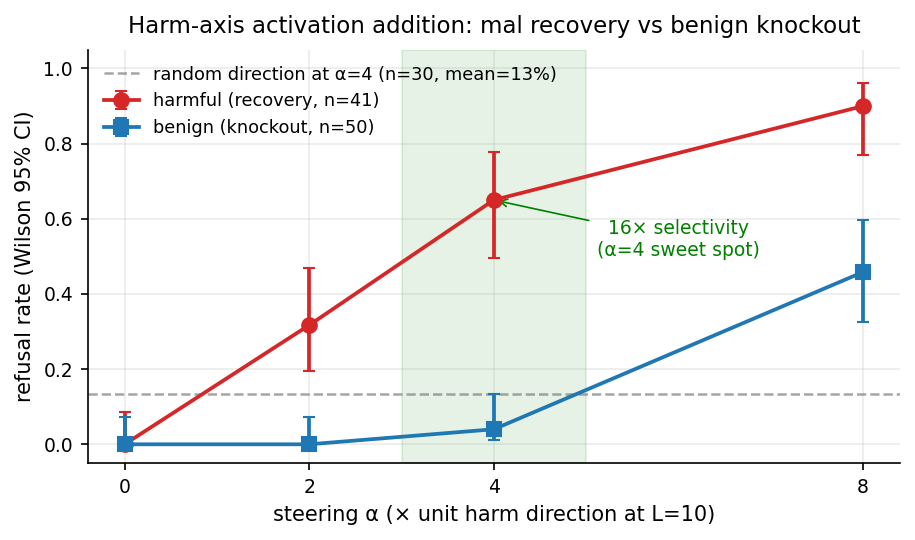}
\hfill
\includegraphics[width=0.49\linewidth]{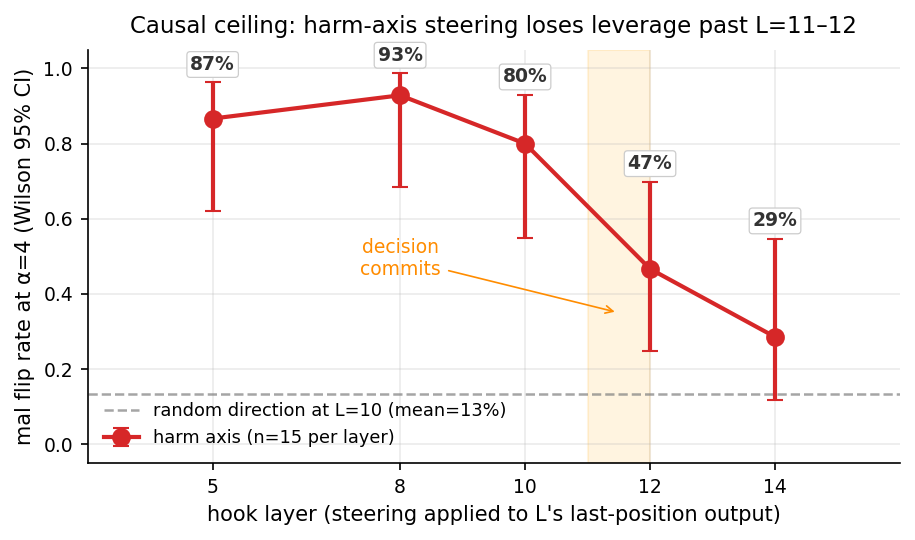}
\caption{\emph{Left}: mal recovery and benign knockout vs.\ steering magnitude $\alpha$
(Wilson 95\% CIs); $\alpha{=}4$ is the selectivity sweet spot. \emph{Right}: mal
recovery at $\alpha{=}4$ vs.\ hook layer; leverage vanishes past $L11$--$12$.}
\label{fig:steer}
\end{figure}

We add $\alpha\vharm$ to the last-position residual at every generation step, hooking
$L10$, on (malicious, comply) and (benign, comply) trajectories from each Indic
language. At $\alpha{=}4$ (${\approx}27\%$ of the natural residual norm) harm-axis
steering recovers refusal in $65\%\,[50,78]$ of malicious-comply trajectories while
refusing only $4\%\,[1,13]$ of benign prompts ($16\times$ selectivity); a matched-norm
random direction flips only ${\sim}13\%$ (${\sim}6\times$ gap). A layer sweep at
$\alpha{=}4$ gives $87\%$ at $L5$, $93\%$ at $L8$, $80\%$ at $L10$, $47\%$ at $L12$,
$29\%$ at $L14$: strong leverage anywhere in $L5$--$10$, dropping sharply past
$L11$--$12$, consistent with the refuse/comply decision committing there. The
output-position refusal axis, although correlationally clean, does not reliably steer
($0$--$1/5$), supporting the upstream harm representation as the causal driver
(\cref{fig:steer}).

\section{MoE expert routing collapses by the output}\label{app:moe}
\begin{figure}[h]
\centering
\includegraphics[width=0.95\linewidth]{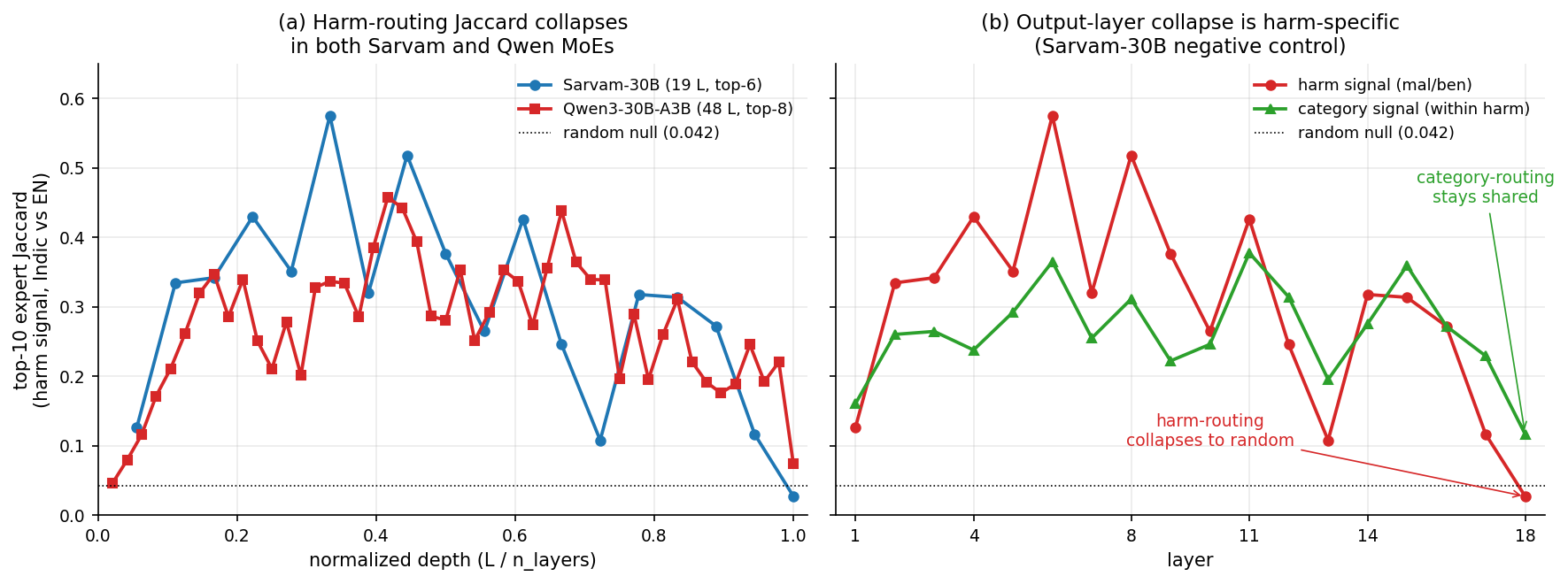}
\caption{Top-$10$ harm-correlated expert Jaccard between Indic languages and English.
\emph{(a)} Both \sarvam{} and \qwenmoe{} peak mid-network and collapse to the random
null at the output. \emph{(b)} \sarvam{} harm signal vs.\ within-harm category signal:
at $L18$ harm is at null ($0.027$) while category is ${\approx}3\times$ above ($0.117$).}
\label{fig:jaccard}
\end{figure}

Per prompt we record a top-$6$ expert-usage fingerprint at each MoE layer and define
each layer's harm-relevant expert set as the top-$10$ experts by
$|\mathrm{Pearson}(\mathrm{usage},\mathrm{harm\_label})|$. The mean cross-lingual
Jaccard rises through the mid-network, peaks at $0.43$ at $L11$ ($+8.7\sigma$ above the
random null $0.041$), and collapses to $0.027$ at $L18$ (\cref{fig:jaccard}). The
collapse is harm-specific: a within-harm \emph{category} signal stays at $0.117$
(${\approx}3\times$ null) at $L18$. The same peak$\to$random shape replicates in
\qwenmoe{}. Concretely, English's top-$10$ harm experts at $L11$ overlap $5$--$7$ of
$10$ with each Indic language; at $L18$ that drops to $0$--$2$. The shared upstream harm
signal is realized through near-disjoint expert sets per language by the output---a
candidate mechanism for why one language's safety routing can fail in isolation.

\section{Reproducibility}\label{app:repro}
\textbf{Data.} Every file named below is attached to this submission as an ancillary
file; \texttt{anc/README.md} maps each to the figure or table it backs.
\textbf{Environment.} \sarvam{} runs under \texttt{transformers==4.57.x}; heavy
intervention/perplexity sweeps ran on a separate GPU box and were synced back.
\textbf{Detection \& steering.} Cross-lingual axes, language probe, and logit-lens from
the upstream extraction pipeline; steering from the activation-addition sweeps
(\texttt{steering\_pilot\_v6.jsonl}, layer sweep, random control). \textbf{Circuit.}
\cref{fig:attr} from direct logit attribution (\texttt{dla.jsonl}) and
gradient$\times$activation (\texttt{grad\_attrib.jsonl}, \texttt{grad\_heads.json});
\cref{tab:map} combining judged recovery sweeps (\texttt{attn18\_eos\_judged},
\texttt{flip\_sweep}, \texttt{doseresponse}) with WikiText perplexity sweeps
(\texttt{perplexity.json}, \texttt{opposer\_damp\_ppl.json}, \texttt{surgical\_ppl.json}).
\textbf{Labels.} Claude Haiku 4.5 LLM-as-judge; regex/LLM agreement $38/39$ non-empty
cases in a stratified sample. \textbf{Cross-model (\qwenmoe{}).} Attribution from
\texttt{qwen\_mechanism.json} (DLA, $\cos$) and \texttt{qwen\_grad.json}; cross-lingual gap
from \texttt{qwen\_labels.json} (scale-free ratios via \texttt{scripts/normalize.py}); lever
generations in \texttt{qwen\_lever.jsonl}, reported qualitatively (recovers-but-saturates).

\end{document}